\pdfoutput=1

\documentclass{article} %
\usepackage{iclr2027_conference,times}

\usepackage{amsmath,amsfonts,bm}

\def\eqref#1{equation~\ref{#1}}

\def\1{\bm{1}}

\DeclareMathAlphabet{\mathsfit}{\encodingdefault}{\sfdefault}{m}{sl}
\SetMathAlphabet{\mathsfit}{bold}{\encodingdefault}{\sfdefault}{bx}{n}

\usepackage{hyperref}
\usepackage{url}
\usepackage{graphicx}
\usepackage{booktabs}
\usepackage{makecell}
\usepackage{multirow}
\usepackage{adjustbox}
\usepackage{nicematrix}
\usepackage{caption}
\usepackage{wrapfig}
\title{Load-Aware Quadruped Navigation with Online Contact-based Geometry Estimation}

\author{
Qi Yin \quad XiaoLi Zhang\quad Yangming Shi \\
 Colorado School of Mines, Golden, CO 80401, USA 
}

\iclrfinalcopy
\begin{document}

\maketitle

\begin{abstract}
Navigating cluttered environments while carrying objects remains a challenging problem for quadruped robots. Compared with works that focused on load-aware locomotion adaptation to maintain gait stability, existing autonomous navigation systems generally ignore the influence of carried objects' geometries and use fixed footprint inflation, leading to overly conservative navigation. To fill this knowledge gap, we propose a two-phase load-aware navigation learning framework. In Phase 1, we train LOADNav, a load-aware navigation policy via reinforcement learning, enabling the robot to adapt its trajectory and clearance behavior according to the geometric size of the carried object. In Phase 2, we train CoGE, an online geometry predictor via imitation learning, which estimates unknown payload geometry from occasional contact events. We evaluate the framework in cluttered environments with loads of varying sizes and shapes. The results show that our navigation policy exhibits clear load-aware behaviors, and the online geometry predictor estimates object geometry from occasional contact events, refining its estimate and adapting the navigation behavior accordingly. Without requiring any prior knowledge of the payload, it narrows the gap to geometry-aware policies. Project website: \url{https://qiii04.github.io/LOADNav/}

\end{abstract}

\begin{figure}[h]
    \vspace{-12pt}
    \centering
    \includegraphics[width=1.0\linewidth]{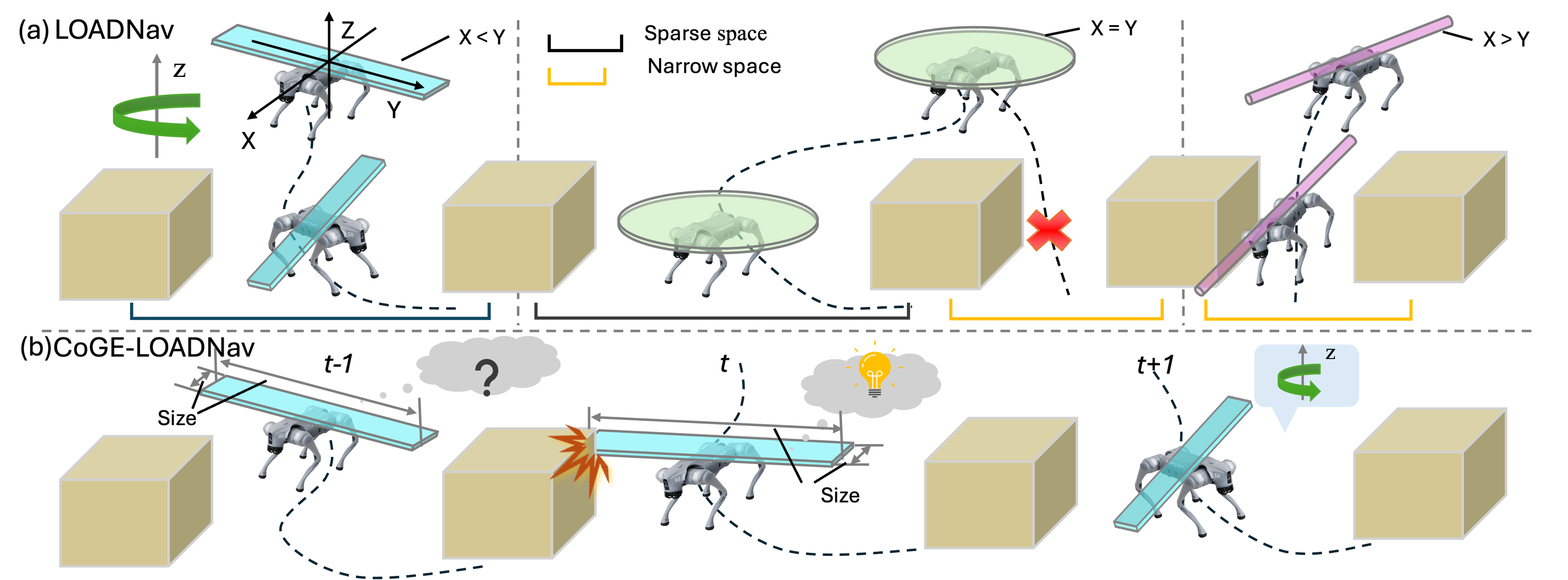}
    \caption{The illustration of showing our proposed two-phase load-aware navigation learning framework. (a) LOADNav (Phase 1): with privileged object size information, the policy learns load-aware navigation strategies, adapting its behavior across sparse and narrow spaces according to different object geometries ($X < Y$, $X = Y$, $X > Y$); (b) CoGE-LOADNav (Phase 2): the robot infers object geometry from occasional contact events, enabling adaptive navigation without size inputs. }
    \label{fig:1}
    \vspace{-12pt}
\end{figure}
\section{Introduction}
\label{sec:1}
Quadruped robots have demonstrated promising potential in object transportation tasks \citep{7758092, unitree_go1, unitree_go2}. Recent research studies have investigated the effect of carried objects on quadruped locomotion, focusing on maintaining stable gaits under varying payload conditions \citep{sombolestan2024adaptive, amanzadeh2024predictive,xu2025facetforceadaptivecontrolimpedance,zhi2025learningunifiedpolicyposition}, and improving controller robustness by modeling or predicting the dynamic properties of the payload \citep{chang2025robustnesslearningunknowndynamic, 10924714}.

However, the impacts on navigation when the carried objects significantly exceed the robot's size remain unexplored. Transporting large objects substantially enlarges and irregularizes the robot's geometric boundary, fundamentally altering the geometric constraints imposed on the navigation system. Although many previous studies have been made in autonomous navigation for quadruped robots \citep{wang2025omni,liu2025compass}, existing approaches typically assume a fixed and compact robot size \citep{fu2022coupling,xu2025navrl,wellhausen2023artplanner}, and therefore struggle to handle the irregular geometry introduced by large carried objects. Moreover, the geometric boundary of the robot varies with the shape and size of the carried object, requiring the navigation system to adapt to diverse geometric constraints. This challenge is further compounded by the fact that the geometry of the payload, mounted on the robot's back, lies in the blind spot of onboard visual and LiDAR sensors. Consequently, the system cannot directly perceive the object's geometry, necessitating inference through proprioceptive feedback.

To address these challenges, we propose a \textbf{two-phase load-aware navigation learning} framework. In Phase 1, we train a navigation policy called \textbf{LOADNav} via reinforcement learning that enabled the robot to avoid obstacles and reach target goals while carrying objects of varying geometries. The policy demonstrated emergent load-aware behaviors. For instance, a board mounted perpendicularly to the robot rotated sideways to pass through narrow passages, circular objects that uniformly enlarged the robot's bounding contour detoured around overly narrow regions, and tubes aligned with the robot's body shape sought the shortest traversable path, as illustrated in Fig. \ref{fig:1}(a). In Phase 2, we train an online geometry predictor called \textbf{CoGE-LOADNav} via imitation learning, enabling the robot to estimate the geometry of the carried object during deployment without prior knowledge, as illustrated in Fig. \ref{fig:1}(b). Results indicate that the predictor effectively estimated object size through occasional contact events. Importantly, our method does not intentionally induce collisions but leverages contact events when they occur to refine its geometry estimate and avoid future collisions. Our main contributions are as follows:  
 \begin{itemize}
    \item We investigate how payload geometry affects navigation behavior and show that geometry-conditioned policies adapt their trajectories and orientations to different payloads.
    \item We analyze how payload geometry is identifiable only through occasional contact events. Unlike dynamic parameters that continuously affect proprioception, payload geometry becomes observable only through contact-induced motion deviations. Based on this property, we train CoGE, a contact-based online geometry estimator.%
    \item We identify forgetting in finite-memory estimators. Once contact information leaves the observation window, the estimator inevitably loses the acquired geometry. We address this with a recurrent estimator that preserves contact information over time.
    \item We validate our framework in simulation and real-world experiments, demonstrating improved navigation performance over baselines and effective online geometry estimation.
 \end{itemize}

\section{Related Works}
\label{sec:2}
\textbf{Load-Aware Locomotion.} Payload carrying significantly affects the locomotion stability and control of quadrupedal robots, and existing work broadly addresses this challenge along three directions: model-based adaptive control \citep{sombolestan2024adaptive, amanzadeh2024predictive, zeng2024adaptive, minniti2021adaptive}, online inertial parameter estimation \citep{10924714, sobanbabu2025samplingbasedidentificationactiveexploration}, and learning-based adaptation \citep{chang2025robustnesslearningunknowndynamic, singh2024panospayloadawarenavigationoffroad}. On the model-based side, \citet{sombolestan2024adaptive} develop adaptive force-based controllers that leverage proprioceptive feedback to compensate for payload-induced disturbances, while \citet{amanzadeh2024predictive} integrate an indirect adaptive law with MPC to enable robust payload transport under terrain and load uncertainty. For online estimation, \citet{10924714} demonstrate that inertial parameters of an unknown payload can be identified in real time on a Unitree Go1 and fed back into a full-dynamics MPC controller, yielding measurable improvements in posture stability and foot tracking. On the learning side, \citet{chang2025robustnesslearningunknowndynamic} train an RL policy capable of inferring and stabilizing unknown dynamic loads using proprioception alone. Despite these advances, existing work focuses predominantly on how payload affects locomotion, while its impact on navigation remains largely unexplored, particularly when the carried object extends beyond the robot's own footprint. Moreover, existing payload estimation methods primarily estimate dynamic properties relevant to locomotion, such as mass and inertia, rather than the payload's geometry required for navigation.

\textbf{Planning-based Navigation.} Navigation algorithms for robots can be categorized into planning-based and learning-based approaches. Planning-based methods explicitly search for paths over a spatial representation of the environment by computing traversability or motion costs \citep{fu2022coupling, ren2025top, scheidemann2025obstacle, wellhausen2023artplanner, wellhausen2021rough, luan2024intelligent, han2025neupan, garg2025objectreactlearningobjectrelativecontrol, Yin_2026_CVPR}, whereas learning-based methods use neural networks to implicitly map perception inputs to navigation commands \citep{xu2025navrl, 10611254, xu2025rebot, yao2025towards, huang2024collision, wang2025omni, liu2025compass, wang2025x, zhang2025motionpriorsreimaginedadapting, 10.22260/ISARC2026/0071}. Existing navigation approaches universally treat robot footprint as a fixed, compact constraint in the algorithm. For instance, VP-Nav \citep{fu2022coupling} computes a distance-based penalty map that keeps the robot at a fixed clearance from obstacles, and ArtPlanner \citep{wellhausen2023artplanner} assigns motion costs based on a fixed bounding volume enclosing the robot. On the learning-based side, NavRL \citep{xu2025navrl} enforces safety by constraining the action space through a fixed safety shield. However, this assumption of a fixed robot footprint breaks down when a carried object enlarges and irregularizes the robot's geometry.

\textbf{Two-phase Learning.} Two-phase learning has been widely adopted to facilitate sim-to-real transfer in quadruped robots \citep{kumar2021rma, miki2022learning, cheng2024extreme, loquercio2023learning, lee2020learning, agarwal2023legged, nahrendra2023dreamwaq, yin2025teacher}. In the first phase, the policy is trained with privileged information to accelerate learning, while in the second phase, imitation learning is employed to distill the policy into a deployable model that estimates the privileged features from limited proprioceptive inputs. In legged locomotion, RMA \citep{kumar2021rma} regresses a latent encoding of environment factors from the robot's state-action history, \citet{miki2022learning} train robots to handle noisy height map inputs through a two-phase framework, and \citet{agarwal2023legged} enable height feature estimation from depth image inputs. Beyond locomotion, this paradigm has also been applied in manipulation \citep{lum2025dextrah, torne2024reconciling, chen2023visual, chen2025clutterdexgrasp, fuchioka2024robotic, yoo2025kinesoftlearningproprioceptivemanipulation}, where DextrAH-G \citep{lum2025dextrah} estimates object pose from depth images. However, in these studies, the privileged information continuously affects the robot's dynamics or perception, such that each step of the proprioceptive history contains information about them. In contrast, the payload geometry considered in our work does not continuously affect proprioception and is revealed only through intermittent contact with the environment.

\section{Method}
\label{sec:3}
\subsection{Problem Formulation}
\label{sec:3.1}
\begin{figure}[h]
    \centering
    \vspace{-10pt}
    \includegraphics[width=1.0\linewidth]{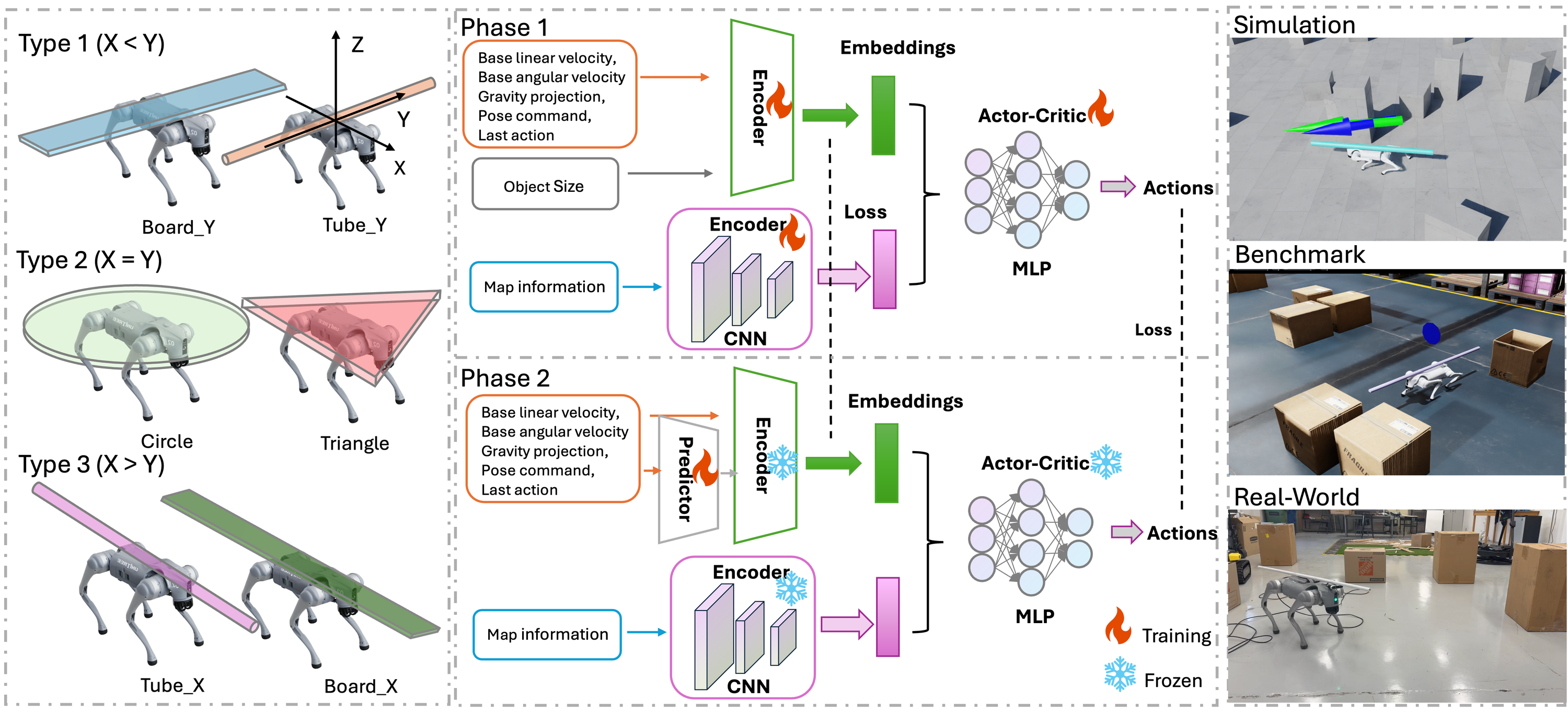}
    \caption{Overview of our proposed navigation framework.  (Left) Three object geometry types categorized by their X and Y dimensions. (Middle) Two-phase training pipeline: Phase 1 trains the load-aware navigation policy with privileged object size input via Reinforcement Learning, while Phase 2 trains the contact-based predictor via Imitation Learning with the main policy network frozen. (Right) Evaluation across simulation, benchmark, and real-world environments.}
    \label{fig:2}
    \vspace{-15pt}
\end{figure}

We consider a quadruped robot navigating from a start position to a goal position while carrying a payload mounted on its back. The navigation task is defined as a 2D point navigation problem: given a  planar (xy) goal position $\mathbf{g} \in \mathbb{R}^2$, the robot must reach $\mathbf{g}$ while avoiding collisions with obstacles in the environment. The payload may change the robot's  planar physical boundary, introducing geometric constraints that vary with the payload's  planar geometric shape and size. We define the robot's local coordinate frame with the heading direction as the x-axis and the lateral direction as the y-axis. The payload is assumed to be rigidly and centrally symmetric attached to the robot's back without relative motion or rotation. The payload geometry is characterized by two parameters:
$\mathbf{s} = (s^x, s^y) \in \mathbb{R}^2$,
where $s_x$ and $s_y$ denote the absolute extent of the payload along the x and y axes in the robot's local frame, respectively. Based on the geometric relationship between the payload and the robot's heading direction, we define six representative payload types, as illustrated in Fig. \ref{fig:2}. These payload types impose distinct geometric constraints on navigation, motivating our proposed two-phase load-aware navigation framework detailed in the following sections.

\subsection{Overview Framework}
\label{sec:3.2}
We propose a two-phase load-aware navigation learning framework, as illustrated in Fig. \ref{fig:2}. The policy network takes as input the robot's proprioceptive states and a local occupancy map constructed from onboard sensors. In the first phase, we train a load-aware navigation policy via reinforcement learning, enabling the robot to acquire distinct navigation strategies for different payload types, where the payload geometry $\mathbf{s}$ is provided as privileged information. In the second phase, to further enhance deployment flexibility, the trained policy network was frozen, and a contact-based online geometry predictor is trained via imitation learning to estimate $\mathbf{s}$ from onboard proprioceptive sensing, enabling load-aware navigation without prior geometric knowledge.

\subsection{Phase 1: Load-Aware Navigation Policy}
\label{sec:3.3}
\textbf{Observation.} The observation space consists of three components: $\mathbf{o} = (\mathbf{o}^p, \mathbf{o}^s, \mathbf{o}^m)$. The proprioceptive state $\mathbf{o}_p$ includes the robot's base linear and angular velocities over the past 5 frames, the current projected gravity vector, the relative goal position $\mathbf{p}_{b}^{g,xy}\in \mathbb{R}^2$, and the previous action. The payload geometry $\mathbf{o}^s = (s^x, s^y) \in \mathbb{R}^2$ encodes the geometric extent of the payload along the robot's local x and y axes. The map observation $\mathbf{o}_m$ is a local occupancy map centered on the robot, spanning $[-2.5, 2.5]$ m along both axes with a resolution of $0.05$ m, resulting in a $100 \times 100$ grid flattened into a $10000$-dimensional vector. The total observation dimension is $10040$.

\textbf{Action.} The action space consists of a 3-dimensional velocity command $\mathbf{a} = (v^x, v^y, \omega) \in \mathbb{R}^3$, representing the desired linear velocities along the x axis, y axis, and the angular velocity. These commands are executed by our pre-trained flat locomotion controller based on the default velocity tracking task from Isaac Lab \citep{isaaclab}.

\textbf{Network Architecture.} The policy network adopts a dual-branch actor-critic architecture. The proprioceptive branch processes $(\mathbf{o}^p, \mathbf{o}^s)$ through a two-layer MLP encoder, while the map branch processes $\mathbf{o}^m$ through a three-layer CNN followed by a linear projection. The outputs of both branches are concatenated and fed into independent actor and critic heads, outputting the velocity command $\mathbf{a}$ and state value, respectively. The entire network is trained using Proximal Policy Optimization (PPO) \citep{schulman2017proximalpolicyoptimizationalgorithms}, with detailed hyperparameters provided in the Appendix Table \ref{tab:4}.

\textbf{Reward Functions.}
The reward function consists of two components: terminal rewards $r_{\text{terminal}}$ triggered by episode termination conditions, and dense rewards $r_{\text{dense}}$ that provide continuous feedback during navigation:
\begin{equation}
r = r_{\text{terminal}} + r_{\text{dense}}
\end{equation}
The terminal rewards are defined as:
\begin{equation}
r_{\text{terminal}} = \lambda_1 r_{\text{succ}} + \lambda_2 r_{\text{timeout}} + \lambda_3 r_{\text{robot}} + \lambda_4 r_{\text{stuck}},
\end{equation}
The dense rewards are defined as:
\begin{equation}
r_{\text{dense}} = \lambda_5 r_{\text{obj}} + \lambda_6 r_{\text{vel}} + \lambda_7 r_{\text{head}} + \lambda_8 r_{\text{step}},
\end{equation}

The terminal rewards encode four termination conditions: a success reward $r_{\text{succ}}$ upon goal reaching, and penalties for timeout $r_{\text{timeout}}$, robot body contact with the environment $r_{\text{robot}}$, and becoming stuck $r_{\text{stuck}}$, which is triggered when both the planar velocity and angular velocity of the robot fall below their respective thresholds within a single timestep. The dense rewards include a velocity projection reward $r_{\text{vel}}$ that encourages the robot to move toward the goal, a heading alignment reward $r_{\text{head}}$ that encourages the robot to move along its forward heading direction, avoiding backward or sideways locomotion, an object collision penalty $r_{\text{obj}}$ that penalizes contacts between the loaded object and the environment, and a per-step time penalty $r_{\text{step}}$ that encourages the robot to complete the task in fewer steps. Detailed formulations and reward weights are provided in the Appendix Section \ref{sec:7.1}.

\textbf{Navigation Curriculum.} We design a Navigation Curriculum to progressively expose the robot to increasingly challenging navigation scenarios. The training terrain is organized as a $10 \times 30$ grid, where each terrain cell contains obstacles at random positions. Obstacle count increases linearly along the row direction, with each row corresponding to a distinct difficulty level. At the beginning of each episode, the robot's initial position and goal are randomly sampled within the current terrain cell, subject to a minimum clearance constraint. Each robot independently maintains its curriculum level $l_i$ within its assigned column. A sliding window of length $N$ tracks the robot's recent navigation trials, and when the window is full, the curriculum level is updated according to:
\begin{equation}
    l_i' = \begin{cases} l_i + 1, & s_i \ge \tau_{\text{succ}} \\ l_i - 1, & s_i \le \tau_{\text{fail}} \\ l_i, & \tau_{\text{fail}} < s_i < \tau_{\text{succ}} \end{cases},
\end{equation}

where $s_i$ is the success rate within the current window, and $\tau_{\text{succ}}$ and $\tau_{\text{fail}}$ are the promotion and demotion thresholds respectively. Following each level transition, a cooldown period is enforced during which the curriculum level remains fixed, preventing boundary oscillation. Detailed curriculum parameters are provided in the Appendix Table \ref{tab:5}.
\subsection{Phase 2: Contact-based Online Geometry Predictor}
\label{sec:3.4}
Before introducing CoGE, we first clarify how the payload geometry $s$ differs from the privileged information estimated in existing two-phase methods, and why this distinction calls for event-driven estimation. Let $q_t$ denote the robot pose, $O$ the obstacle set, and 
$B(q_t,s)=q_t\oplus(B_0\cup F(s))$ the effective footprint formed by 
the robot body $B_0$ and payload $F(s)$. Contact introduces a disturbance 
to the nominal locomotion, such that 
$q_{t+1}=f(q_t,a_t)+d_t$, where 
$d_t\neq0 \iff B(q_t,s)\cap O\neq\emptyset$.
We assume that the payload is rigidly attached, has negligible dynamic 
effects (mass fixed at $0.1\,\mathrm{kg}$), and lies outside the sensor 
field of view; thus, $o_t^m$ is independent of $s$.

\textbf{Proposition.}
At any contact-free time step $t$, for arbitrary $s$ and $s'$,
$p(o_t^p\mid h_{t-1},a_{t-1},s)
= p(o_t^p\mid h_{t-1},a_{t-1},s')$.
Hence, the Fisher information of $o_t^p$ about $s$ is zero,
i.e., $\mathcal{I}_t(s)=0$.

\textbf{Proof.}
Without contact, $d_t=0$, making the transition dynamics independent of $s$.
Since $o_t^p$ consists of velocity, projected gravity, goal command, and the previous action, it is also independent of $s$.
Therefore, the proposition holds.
Consequently, the cumulative information about $s$ along a trajectory satisfies
\begin{equation}
\sum_t \mathcal{I}_t(s)
=
\sum_{t\in\mathcal{T}_c}\mathcal{I}_t(s),
\label{eq:cumulative_information}
\end{equation}
where $\mathcal{T}_c$ denotes the set of contact times.
Information about $s$ therefore increases only upon contact.
Before the first contact along an axis, its estimation error cannot improve beyond the prior variance, e.g.,
\begin{equation}
\mathbb{E}\left[(\hat{s}_x-s_x)^2\right]
\geq \mathrm{Var}(s_x).
\label{eq:prior_variance}
\end{equation}

In existing two-phase methods, privileged information $\xi$ continuously affects the transition dynamics $f(q_t,a_t,\xi)$, providing information at every time step. In contrast, information about $s$ arises only from contact events. The geometry estimator must therefore detect contact and update accordingly. To enable deployment flexibility without requiring prior knowledge of the object geometry, we introduce a contact-based online geometry predictor CoGE that estimates the object size $\hat{o}^s_t$ from occasional contact events. The predicted geometry $\hat{o}^s_t$ is then concatenated with the proprioceptive state and fed into the frozen policy network to produce the action $\hat{a}_t = \pi_{\theta}(o^p_t | \hat{o}^s_t | o^m_t)$.
The predictor is trained via imitation learning with the Phase 1 load-aware navigation policy $ \pi_{\theta}$ frozen. Two MSE losses are jointly optimized $\mathcal{L} = \mathcal{L}_{\text{obj}} + \mathcal{L}_{a}$: a geometry loss $\mathcal{L}_{\text{obj}} = |\hat{o}^s_t - o^s_t|^2$ that supervises the predicted object size against the ground truth, and an action loss $\mathcal{L}_{a}  = |\hat{a}_t - a_t|^2$ that encourages the Phase 2 policy to reproduce actions of Phase 1 $a_t = \pi_{\theta}(o^p_t | o^s_t | o^m_t)$. The predictor is optimized using the Adam optimizer, with all other network weights frozen.
\section{Experimental Results}
\label{sec:4}
\subsection{Experimental Setting}
\label{sec:4.1}
All policies are trained in NVIDIA Isaac Lab using 512 parallel Unitree Go2 robots on a single RTX 4090 GPU. Phase 1 trains the load-aware navigation policy on a $10 \times 30$ terrain grid with the Navigation Curriculum for 3000 episodes, requiring approximately 10 hours. Phase 2 trains the contact-based online geometry predictor on a $10 \times 10$ terrain grid with fixed obstacle density of 80 for 1000 episodes, requiring approximately 4 hours. The loaded object is rigidly attached to the robot and comprises six shape categories with randomly sampled dimensions; detailed size ranges are provided in the Appendix Table \ref{tab:6}. The robot's linear velocity was clipped to $[-1.5, 1.5]$ m/s and angular velocity to $[-1.0, 1.0]$ rad/s. We evaluated all methods using two metrics: 1) \textbf{Success Rate (SR)}, defined as the percentage of episodes in which the robot reaches within 0.5m of the goal; 2) \textbf{Average Collision Times (ACT)}, measuring the average number of collisions involving both the robot body and the loaded object per successful trial. We compared against the following methods: 1) VP-Nav \citep{fu2022coupling}: A planning-based navigation policy with occupancy map inflation
and the Obstacle Distance Map design. 2)  VP-Nav-I (inflated): the same planner with the inflation radius enlarged to cover the payload. 3) NavRL \citep{xu2025navrl}: A reinforcement learning-based navigation policy with safety rewards and Policy Action Safety Shield. 4) LOADNav (with given object size): Our load-aware policy that requires object size as input. 5) LOADNav-B (size-blind): LOADNav trained without payload geometry input.  6) CoGE-LOADNav: Our full method equipped with a contact-based online geometry predictor with finite memory. 7) CoGE-LOADNav+ (optimized): an improved variant with recurrent temporal training. %

\subsection{Evaluation in the Simulator}
\label{sec:4.2}
\subsubsection{Load-aware Navigation 
Performance}
\label{sec:4.2.1}
This experiment serves two purposes: (1) to show that LOADNav exhibits load-aware navigation behavior, adapting its strategy to the geometry of the carried payload, and (2) to verify that LOADNav performs consistently across payload types and obstacle densities, and that CoGE-LOADNav approaches this performance without requiring the geometry input.

\begin{figure}[h]
    \centering
    \includegraphics[width=1.0\linewidth]{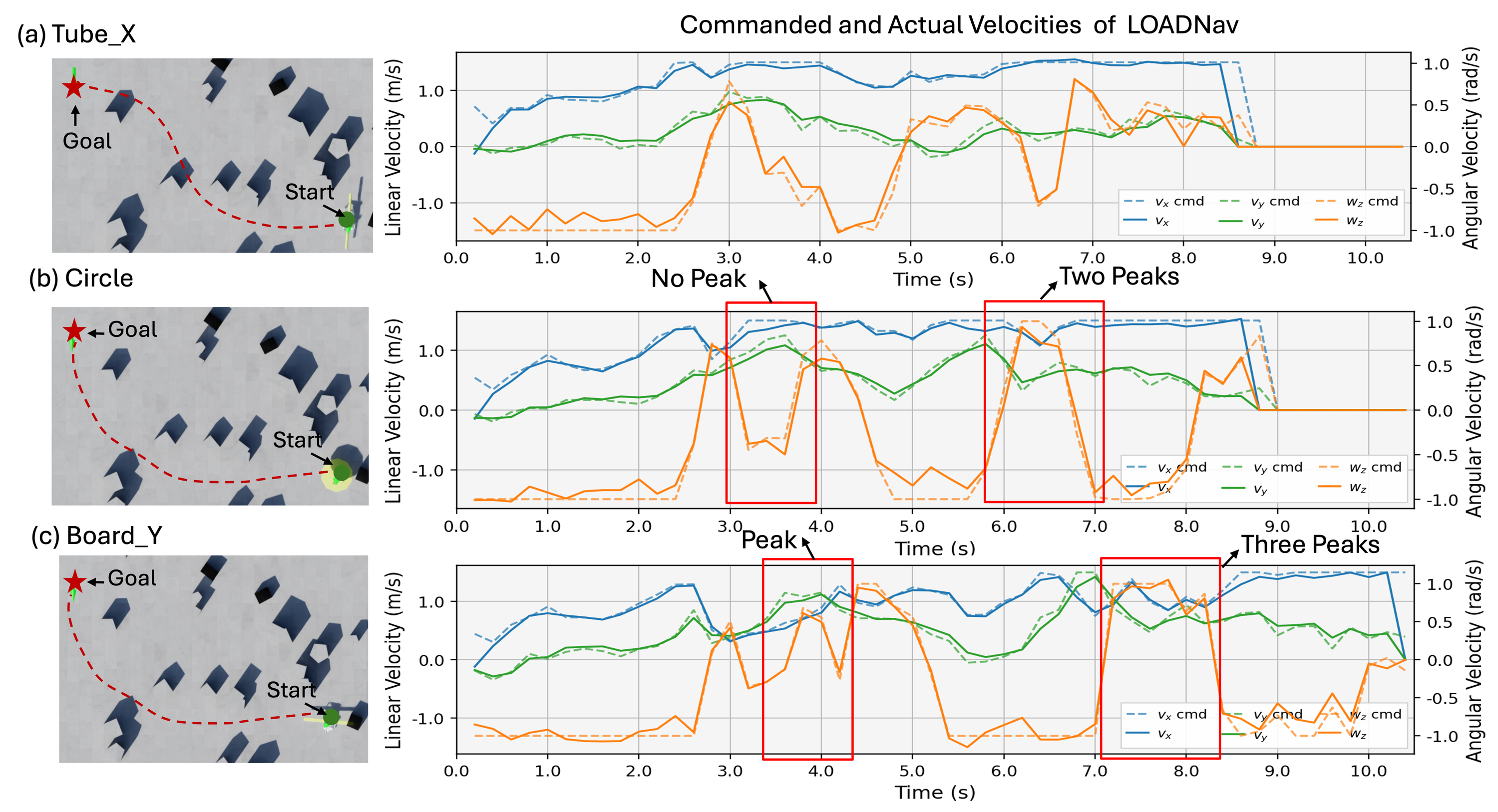}
    \caption{LOADNav trajectories and velocity profiles for different carried objects (N=80). (a) Tube\_X (Length = 1.6 m). (b) Circle (Radius = 0.5 m). (c) Board\_Y (Length = 1.6 m).}
    \label{fig:3}
    \vspace{-10pt}
\end{figure}

\textbf{Load-aware behavior.}
The trajectories and velocity profiles in Fig.~\ref{fig:3} demonstrate that LOADNav exhibits load-aware navigation behavior. Given the same environment, start, and goal, the policy adapts its strategy according to the geometry of the carried payload. For Tube\_X, which is aligned with the robot's heading, the robot selects the shortest path through relatively narrow passages, as the elongated payload imposes minimal lateral clearance constraints. For Circle, whose extent is uniform in all directions, the robot avoids narrow spaces and takes a wider detour. When carrying Board\_Y, oriented perpendicular to the heading, the robot likewise detours; however, the velocity profile shows an additional angular-velocity peak when passing between obstacles, indicating that the robot actively rotates its body to avoid collision and prevent the board from getting stuck.

\textbf{Performance across payload types and densities.}
We evaluate LOADNav and CoGE-LOADNav across six payload types and their mixture under obstacle densities from $N=20$ to $N=80$ on a $5\times5$ terrain grid with 50 robots over 200 episodes; full results are given in Appendix Table~\ref{tab:7}. LOADNav achieves an average success rate of \textbf{91.26\%} across all settings and remains robust under dense obstacles. Without any geometry input, CoGE-LOADNav achieves \textbf{89.74\%}, within 1.6 percentage points of LOADNav; its higher ACT reflects the predictor's reliance on contact feedback and is analyzed in Section~\ref{sec:4.2.2}. 

\subsubsection{Contact-based Online Geometry Estimation Analysis}
\label{sec:4.2.2}
\begin{figure}[h]
\vspace{-5pt}
    \centering    \includegraphics[width=1.0\linewidth]{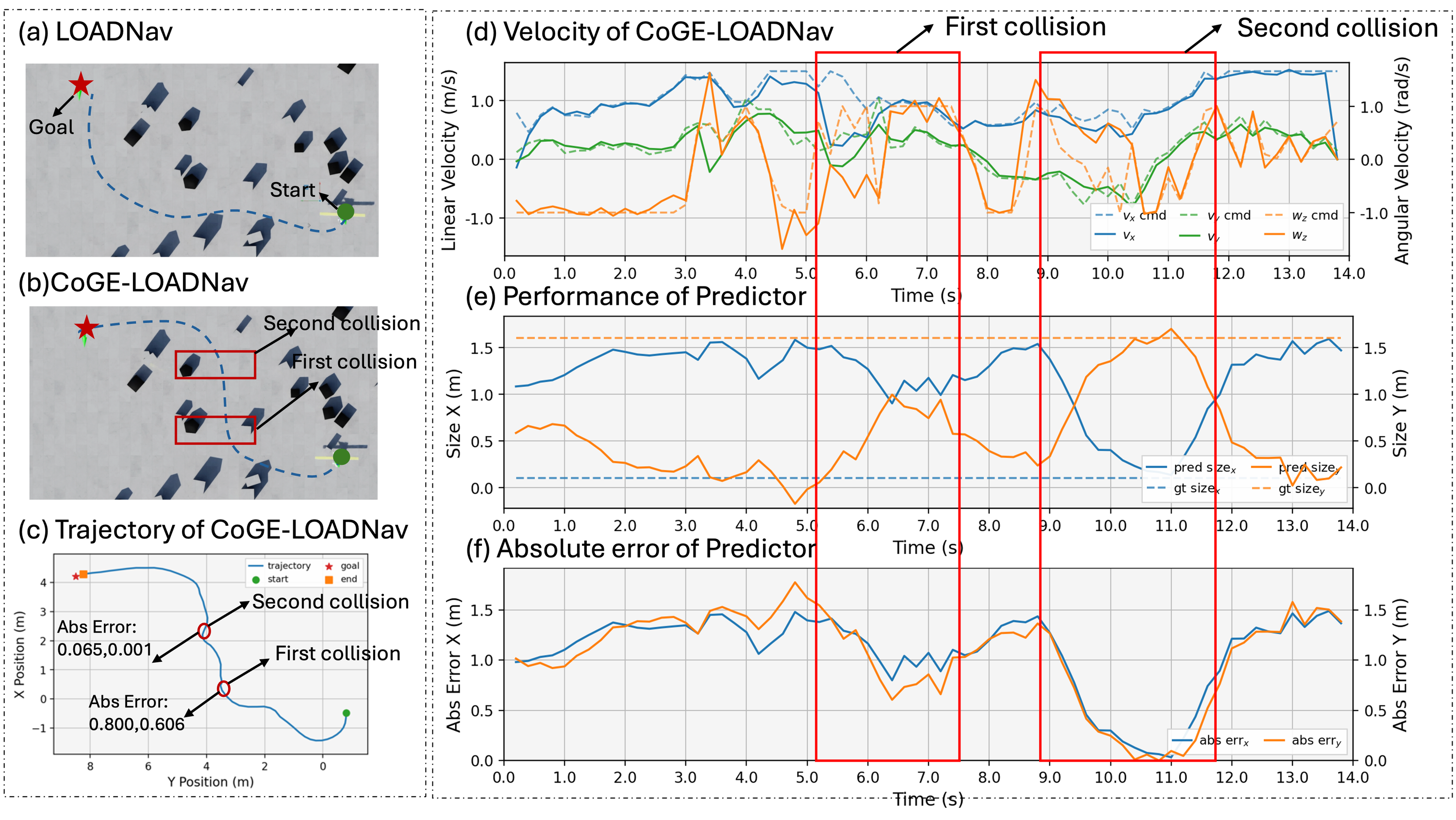}
    \caption{Visualization of LOADNav and CoGE-LOADNav in simulation, illustrating how CoGE-LOADNav infers geometry through collisions. (a-c) Navigation trajectories of LOADNav and CoGE-LOADNav. (d) Commanded and actual velocities of CoGE-LOADNav. (e-f) Predicted object size and absolute errors. }
    \label{fig:4}
    \vspace{-15pt}
\end{figure}

Figure \ref{fig:4} illustrates the geometry estimation process of CoGE-LOADNav on a representative episode. Initially, the predictor has no prior knowledge of the board geometry in the perpendicular direction, resulting in large estimation errors. Upon each contact event, the predictor receives contact feedback and progressively refines its size estimates, with errors dropping significantly after each contact event, as shown in Fig. \ref{fig:4}(f). With improved geometry estimates, the policy is able to make more informed decisions, such as escaping stuck situations and planning appropriate paths and load-aware actions suited to the identified payload shape. The predictor identifies contact events by detecting discrepancies between commanded and actual velocities, as shown in Fig. \ref{fig:4}(d).
\subsection{Memory-enhanced estimation}
\label{sec:4.3}
\begin{figure}[h]
    \vspace{-10pt}
    \centering    \includegraphics[width=1.0\linewidth]{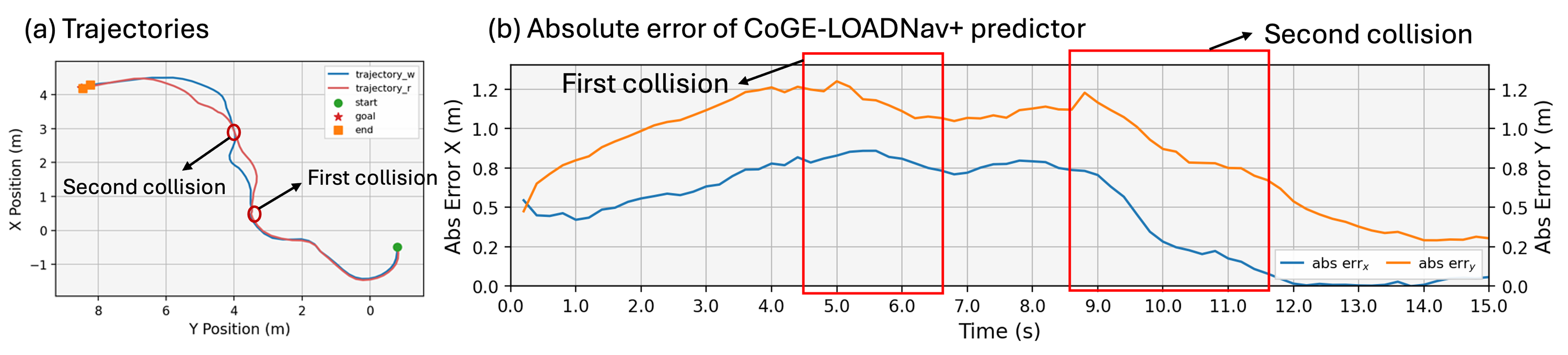}
    \caption{Memory-enhanced geometry estimation with CoGE-LOADNav+. (a) Trajectories of CoGE-LOADNav and CoGE-LOADNav+. (b) Absolute estimation error of the CoGE-LOADNav+.}
    \label{fig:5}
    \vspace{-10pt}
\end{figure}

As shown in Fig. \ref{fig:4}(e)(f), the estimation error decreases rapidly after contact but increases again after approximately $K=5$ steps (1~s), corresponding to the predictor's input window length. This indicates that the contact information is forgotten as it moves out of the window. This behavior is not simply due to insufficient training, but reflects an inherent limitation of finite-window estimation: when the window contains no contact, the optimal window estimator
$\hat{\mathbf{s}}_t^* = \mathbb{E}[\mathbf{s} \mid \mathbf{o}^p_{t-K+1:t}]$
collapses to a constant independent of $\mathbf{s}$, since the window carries no evidence to condition on. Therefore, contact-driven geometry estimation does not satisfy $K$-step decodability for any finite $K$~\citep{efroni2022provable}, and the estimator tends to return toward its prior once the contact signal leaves the window. To retain contact information over longer horizons, the estimator must maintain a state across steps. We therefore train a recurrent memory model whose hidden state is carried across steps within each episode. This temporal training reduces the average size prediction MSE by 15.67\% and the action imitation MSE by 25.09\%. As shown in Fig.~\ref{fig:5}, after two contact events, the estimation error is corrected and remains stable rather than returning to its pre-contact level.
\subsection{Evaluation in the Benchmark}
\label{sec:4.4}
\begin{wraptable}{r}{0.56\linewidth}
    \vspace{-12pt}
    \centering
    \caption{Navigation Performance in the Benchmark.}
    \label{tab:1}

    \scriptsize
    \setlength{\tabcolsep}{1.6pt}
    \renewcommand{\arraystretch}{1.35}

    \begin{tabular}{l cc cc cc cc}
    \toprule
    Method
    & \multicolumn{2}{c}{Average}
    & \multicolumn{2}{c}{Board\_Y}
    & \multicolumn{2}{c}{Circle}
    & \multicolumn{2}{c}{Tube\_X} \\
    \cmidrule(lr){2-3}
    \cmidrule(lr){4-5}
    \cmidrule(lr){6-7}
    \cmidrule(lr){8-9}
    & SR$\uparrow$ & ACT$\downarrow$
    & SR$\uparrow$ & ACT$\downarrow$
    & SR$\uparrow$ & ACT$\downarrow$
    & SR$\uparrow$ & ACT$\downarrow$ \\
    \midrule
    NavRL
    & 30.00 & 0.81
    & 30.00 & \underline{0.33}
    & 50.00 & 0.60
    & 10.00 & 1.50 \\
    VP-Nav
    & 41.67 & 1.20
    & 20.00 & 1.50
    & 55.00 & 0.91
    & 50.00 & 1.20 \\
    VP-Nav-I
    & 28.33 & 0.46
    & 20.00 & 0.50
    & 45.00 & 0.89
    & 20.00 & \textbf{0.00} \\
    \textbf{CoGE-LOADNav}
    & 86.67 & \underline{0.39}
    & \underline{80.00} & 0.50
    & \textbf{90.00} & 0.56
    & \underline{90.00} & \underline{0.11} \\
    CoGE-LOADNav+
    & \underline{88.33} & 0.60
    & 75.00 & 0.93
    & \textbf{90.00} & 0.56
    & \textbf{100.00} & 0.30 \\
    LOADNav-B
    & 63.33 & 0.67
    & 45.00 & 0.89
    & \underline{80.00} & \underline{0.50}
    & 65.00 & 0.62 \\
    \textbf{LOADNav}
    & \textbf{91.67} & \textbf{0.32}
    & \textbf{85.00} & \textbf{0.24}
    & \textbf{90.00} & \textbf{0.11}
    & \textbf{100.00} & 0.60 \\
    \bottomrule
    \end{tabular}
    \vspace{-10pt}
\end{wraptable}
To evaluate navigation performance, we construct a simulated warehouse environment featuring shelves and 100 boxes, as shown in Fig. \ref{fig:10}. %
Real-time occupancy maps are built using RTABMap \citep{labbe2019rtab} from RGB-D images. Each method is evaluated over 20 trials per object category. As shown in Table~\ref{tab:1}, LOADNav achieves the highest average SR of 91.67\%, compared with 41.67\% for VP-Nav and 30.00\% for NavRL. Increasing the inflation radius in VP-Nav-I does not improve performance, with its SR dropping to 28.33\% as few traversable paths remain. CoGE-LOADNav achieves an SR of 86.67\%, while the optimized CoGE-LOADNav+ further improves the SR to 88.33\%. In comparison, LOADNav-B, which does not use payload geometry, achieves only 63.33\% SR, with the largest drop on Board\_Y (85.00\% to 45.00\%), due to its limited geometry-aware navigation behavior. %
The higher ACT of CoGE-LOADNav+ results from a few trials where it recovered from collisions, whereas CoGE-LOADNav became stuck, with collisions in failed trials excluded from ACT. %

\subsection{Real-World Deployment}
\label{sec:4.5}
\begin{figure}[h]
    \vspace{-8pt}
    \centering    \includegraphics[width=1.0
    \linewidth]{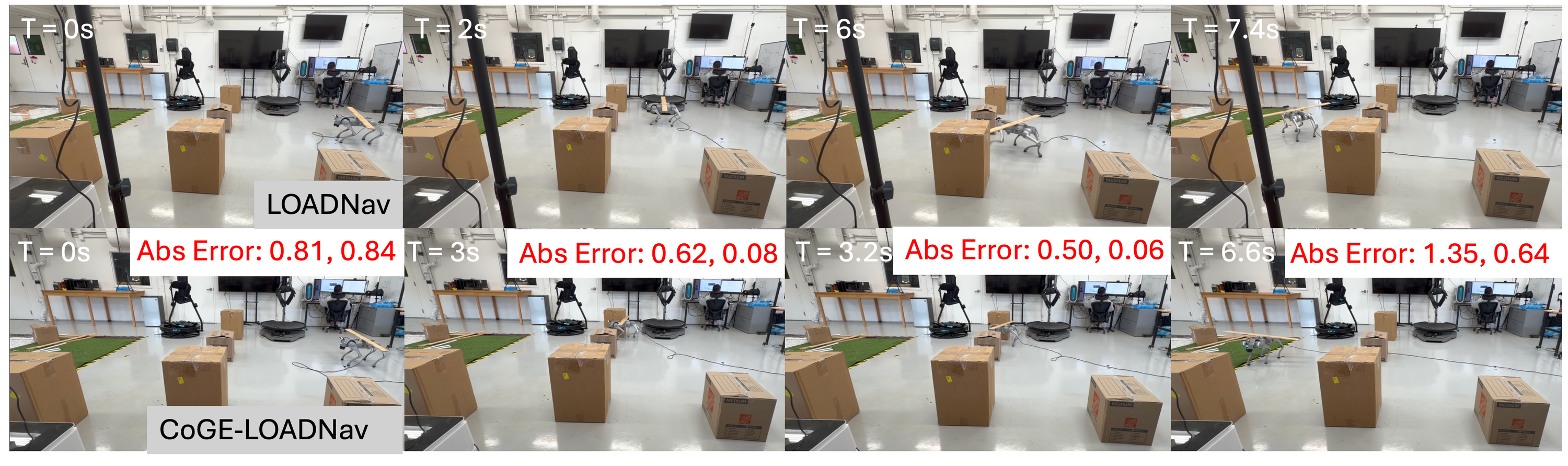}
    \caption{Real-world examples of LOADNav and CoGE-LOADNav.}
    \label{fig:6}
    \vspace{-10pt}
\end{figure}

\begin{wraptable}{r}{0.5\linewidth}
    \vspace{-12pt}
    \centering
    \caption{Navigation Performance in Real-World.}
    \label{tab:2}

    \tiny
    \setlength{\tabcolsep}{1.4pt}
    \renewcommand{\arraystretch}{1.5}

    \begin{tabular}{lcccccccc}
    \toprule
    Method
    & \multicolumn{2}{c}{Average}
    & \multicolumn{2}{c}{Board\_Y}
    & \multicolumn{2}{c}{Tube\_X}
    & \multicolumn{2}{c}{Tube\_Y} \\
    \cmidrule(lr){2-3}
    \cmidrule(lr){4-5}
    \cmidrule(lr){6-7}
    \cmidrule(lr){8-9}
    & SR$\uparrow$ & ACT$\downarrow$
    & SR$\uparrow$ & ACT$\downarrow$
    & SR$\uparrow$ & ACT$\downarrow$
    & SR$\uparrow$ & ACT$\downarrow$ \\
    \midrule
    VP-Nav
    & 45.00 & 1.23
    & 30.00 & 2.00
    & 50.00 & 0.60
    & 55.00 & 1.09 \\
    \textbf{CoGE-LOADNav}
    & 81.67 & \underline{0.33}
    & 80.00 & \underline{0.25}
    & 80.00 & 0.38
    & 85.00 & \underline{0.35} \\
    CoGE-LOADNav+
    & \underline{90.00} & 0.39
    & \underline{85.00} & 0.35
    & \underline{95.00} & \underline{0.37}
    & \underline{90.00} & 0.44 \\
    \textbf{LOADNav}
    & \textbf{98.33} & \textbf{0.04}
    & \textbf{95.00} & \textbf{0.11}
    & \textbf{100.00} & \textbf{0.00}
    & \textbf{100.00} & \textbf{0.00} \\
    \bottomrule
    \end{tabular}
    \vspace{-10pt}
\end{wraptable}
To validate the proposed framework in real-world settings, we deploy both LOADNav and CoGE-LOADNav on the Unitree Go2 platform equipped with a default L1 LiDAR providing a local occupancy map. We evaluate the framework in three scenarios: carrying a 0.03×1.2 m board aligned along the Y-axis, a 1.2×0.09 m tube aligned along the X-axis and a 1.0×0.03 m tube aligned along the Y-axis  in cluttered environments. Each scenario is evaluated over 20 trials. As shown in Table~\ref{tab:2}, LOADNav achieves the highest average SR of 98.33\%, while CoGE-LOADNav achieves 81.67\% without prior knowledge of the payload geometry. The optimized CoGE-LOADNav+ further improves the SR to 90.00\%, reducing the gap to LOADNav. In comparison, VP-Nav achieves only 45.00\% SR.
As demonstrated in Fig. \ref{fig:6}, when carrying the board aligned along the Y-axis with known geometry, the robot exhibits clear load-aware behaviors, such as rotating sideways to pass through relatively open areas. When deploying CoGE-LOADNav without prior knowledge of the payload geometry, the robot may experience slight collisions during early navigation; however, the online geometry predictor rapidly reduces estimation error and corrects its geometry estimate, ultimately guiding the robot to successfully reach the goal.

\subsection{Ablation Studies}
\label{sec:4.6}
\begin{figure}[h]
    \vspace{-10pt}
    \centering    \includegraphics[width=1.0\linewidth]{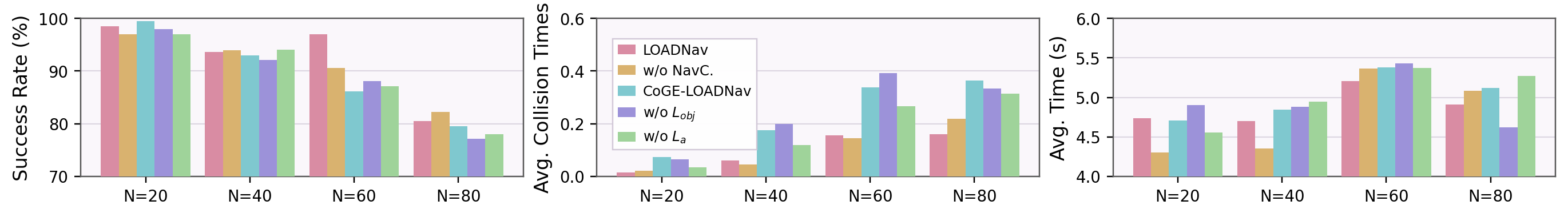}
    \caption{The results of ablation studies comparing LOADNav, CoGE-LOADNav, w/o NavC., w/o $\mathcal{L}_{obj}$, and w/o $\mathcal{L}_a$, in terms of success rate, average collision times, and average time.}
    \label{fig:7}
    \vspace{-10pt}
\end{figure}

To further evaluate the contribution of each component, we conduct ablation studies in simulation under three settings: w/o NavC., where the navigation curriculum is replaced by randomly sampled terrains with random reset positions; w/o $\mathcal{L}{obj}$, where Phase 2 only uses $\mathcal{L}{a}$; and w/o $\mathcal{L}{a}$, where Phase 2 only uses $\mathcal{L}{obj}$. As shown in Fig. \ref{fig:7}, the navigation curriculum significantly improves performance in dense environments, and both loss terms are essential for effective online geometry estimation and navigation.
\section{Discussion and Limitations}
\label{sec:5}
In this paper, we propose a two-phase load-aware navigation learning framework that enables quadruped robots to achieve autonomous navigation in cluttered environments while carrying objects of varying shapes and sizes. Phase 1 trains a load-aware navigation policy that adapts its behavior according to the spatial geometry of the carried object, while Phase 2 enables online estimation of unknown payload geometry  through occasional contact events feedback. A temporal training optimization further mitigates geometry estimation forgetting and reduces the performance gap caused by estimation errors. Extensive experiments in both simulation and real-world environments demonstrate clear load-aware navigation behaviors, such as sideways rotation detour planning, and the effectiveness of the online geometry estimation.

\textbf{Limitations.} Despite the promising results, there are still some limitations. %
First, real-world deployments reveal that uneven terrain negatively affects navigation performance. In particular, on surfaces such as grass and sand, the policy fails to accurately perceive the robot's state, resulting in persistent oscillatory behavior, as the policy is trained exclusively on flat terrain in simulation, causing out-of-distribution effects on velocity tracking. Future work will incorporate uneven terrain during training to enhance the robustness of the proposed framework \citep{10611254}.
Second, while LOADNav, CoGE-LOADNav and CoGE-LOADNav+ achieve comparable performance in simulation, CoGE-LOADNav and CoGE-LOADNav+'s success rate does drop in the real world. Based on our observations, this drop is primarily caused by the gap in collision-force resolution between the simulator and the real world, which degrades the performance of  online geometry predictor. In our future work, we will strengthen sim-to-real transfer to improve performance in the real world.
\clearpage

\subsection*{AI use statement}
In this work, we used generative AI tools only for editing the manuscript to improve readability and identifying relevant literature. All AI-suggested edits and references were reviewed and verified by the authors. We take responsibility for the final content of this work.

\bibliographystyle{iclr2027_conference}
\bibliography{main}
\clearpage
\appendix
\section{Appendix}
\label{sec:7}
\subsection{Reward Functions}
\label{sec:7.1}
The terminal rewards encode four episode termination conditions. A positive reward $r_{\text{succ}}$ is granted upon successful goal reaching, defined as:
\begin{equation}
    r_{\text{succ}} = \mathbb{I}\left(\lVert \mathbf{p}_{b}^{g,xy}\rVert_2 \le \delta\right)
\end{equation}

where $\mathbf{p}_{b}^{g}$ is the relative goal position in the robot body frame and $\delta$ is the goal tolerance. The remaining three terminal rewards impose penalties for undesired terminations: $r_{\text{timeout}}$ penalizes the robot for failing to reach the goal within the maximum episode steps; $r_{\text{robot}}$ penalizes physical contact between the robot body and the environment; $r_{\text{stuck}}$ penalizes the robot when it becomes immobile, defined as:
\begin{equation}
    r_{\text{stuck}} = \mathbb{I}\left(\lVert \mathbf{v}_{b}^{xy}\rVert_2 < v_{\text{th}} \wedge |\omega_b^z| < \omega_{\text{th}}\right)
\end{equation}

where $v_{\text{th}}$ and $\omega_{\text{th}}$ are the velocity thresholds below which the robot is considered stuck.
The dense rewards provide continuous feedback throughout the navigation episode. $r_{\text{vel}}$ encourages the robot to move toward the goal by rewarding the projection of the robot's planar velocity onto the goal direction:
\begin{equation}
    r_{\text{vel}} = \mathbf{v}^{xy}_b \cdot \frac{\mathbf{p}^{g,xy}_b}{\lVert \mathbf{p}^{g,xy}_b\rVert_2+\varepsilon}
\end{equation}

where $\mathbf{v}^{xy}_b$ is the robot's planar velocity. $r_{\text{head}}$ encourages the robot to move along its heading direction by rewarding the cosine similarity between the forward axis and the current velocity direction:
\begin{equation}
    r_{\text{head}} = \begin{cases} \dfrac{v^x_b}{\lVert \mathbf{v}^{xy}_b\rVert_2+\varepsilon}, & \lVert \mathbf{v}^{xy}_b\rVert_2 > 10^{-3}\\[6pt] 0, & \text{otherwise} \end{cases}
\end{equation}

where the zero branch avoids meaningless heading estimates when the robot is nearly stationary. $r_{\text{obj}}$ penalizes collisions between the loaded object and the environment. $r_{\text{step}}$ applies a constant penalty at every timestep, encouraging the robot to complete the task in fewer steps. The reward weights are provided in Table \ref{tab:3}.
\begin{table}[h]
\centering
\caption{Reward Weights.}
\label{tab:3}

\setlength{\tabcolsep}{8pt}
\renewcommand{\arraystretch}{1.05}

\begin{tabular}{lll}

\toprule

Category & Reward Term & Weight \\

\midrule

\multirow{4}{*}{Termination}
& Success reward $r_{\text{succ}}$               & 400.0 \\
& Timeout penalty $r_{\text{timeout}}$          & -40.0 \\
& Robot body contact penalty $r_{\text{robot}}$ & -10.0 \\
& Stuck penalty $r_{\text{stuck}}$              & -50.0 \\

\midrule

\multirow{4}{*}{Dense}
& Velocity projection reward $r_{\text{vel}}$   & 6.0 \\
& Heading alignment reward $r_{\text{head}}$    & 3.0 \\
& Object collision penalty $r_{\text{obj}}$     & -30.0 \\
& Per-step time penalty $r_{\text{step}}$       & -0.5 \\

\bottomrule
\end{tabular}
\end{table}

The PPO hyperparameters used in Phase 1 training are summarized in Table \ref{tab:4}.
\begin{table}[t]
\centering
\caption{PPO Hyperparameters.}
\label{tab:4}

\setlength{\tabcolsep}{10pt}
\renewcommand{\arraystretch}{1.05}

\begin{tabular}{lc}

\toprule

Parameter & Value \\

\midrule

PPO clip parameter                & 0.2 \\
GAE-$\lambda$                     & 0.95 \\
Reward discount factor            & 0.99 \\
Learning rate                     & $3 \times 10^{-4}$ \\
Learning rate schedule            & adaptive \\
Value loss coefficient            & 1.0 \\
Use clipped value loss            & True \\
Entropy coefficient               & 0.003 \\
Desired KL divergence             & 0.01 \\
Max gradient norm                 & 1.0 \\
Number of environments            & 512 \\
Number of env steps per batch     & 24 \\
Learning epochs per batch         & 4 \\
Number of mini batches per epoch  & 8 \\

\bottomrule

\end{tabular}

\end{table}
\subsection{Navigation Curriculum}
\label{sec:7.2}
\begin{figure}[h]
    \centering
    \includegraphics[width=1.0\linewidth]{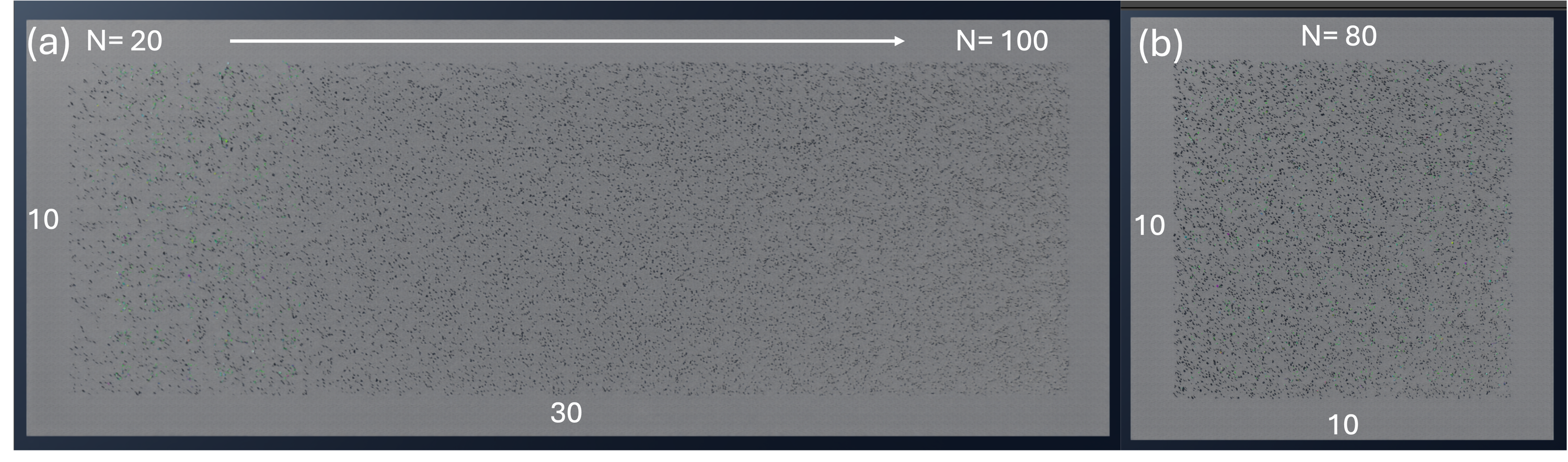}
    \caption{Training environments. (a) Phase 1 with navigation curriculum (N = 20 to 100). (b) Phase 2 without navigation curriculum (N = 80).}
    \label{fig:8}
\end{figure}

The terrain and curriculum parameters adopted in the two-phase training are summarized in Table \ref{tab:5}. The training environments are illustrated in Figure \ref{fig:8}.
\begin{table}[h]
\centering
\caption{Navigation Curriculum Parameters.}
\label{tab:5}
\setlength{\tabcolsep}{8pt}
\renewcommand{\arraystretch}{1.05}

\begin{tabular}{lll}

\toprule

Category & Parameter & Value \\

\midrule

Terrain
& Size & $16 \times 16$ \\

\midrule

\multirow{3}{*}{Box obstacles}
& Length range & $0.25 \sim 0.55$ \\
& Width range  & $0.25 \sim 0.55$ \\
& Height range & $0.4 \sim 1.0$ \\

\midrule

\multirow{2}{*}{Cylinder obstacles}
& Radius range & $0.15 \sim 0.35$ \\
& Height range & $0.4 \sim 1.0$ \\

\midrule

\multirow{4}{*}{Phase 1}
& Number rows         & 30 \\
& Number columns      & 10 \\
& Number objects start & 20 \\
& Number objects end   & 100 \\

\midrule

\multirow{3}{*}{Phase 2}
& Number rows    & 10 \\
& Number columns & 10 \\
& Number objects & 80 \\

\midrule

\multirow{4}{*}{Curriculum window}
& Size         & 10 \\
& Success rate & 0.8 \\
& Fail rate    & 0.5 \\
& Cooldown     & 20 \\

\bottomrule

\end{tabular}

\end{table}
\subsection{Loaded Objects}
\label{sec:7.3}
During training, each robot is assigned a randomly sampled carried object rigidly attached to the base joint. Six object categories are used, each with randomized dimensions sampled from the ranges specified in Table \ref{tab:6}. All six categories are sampled with equal probability.
\begin{table}[t]
\centering
\caption{Object Parameters.}
\label{tab:6}

\setlength{\tabcolsep}{8pt}
\renewcommand{\arraystretch}{1.05}

\begin{tabular}{lll}

\toprule

Object & Parameter & Value \\

\midrule

\multirow{3}{*}{Board\_Y}
& Length range & $0.40 \sim 1.60$ \\
& Width range  & $0.04 \sim 0.20$ \\
& Thickness    & 0.30 \\

\midrule

\multirow{2}{*}{Tube\_Y}
& Length range & $0.40 \sim 1.60$ \\
& Radius range & $0.01 \sim 0.08$ \\

\midrule

\multirow{2}{*}{Circle}
& Radius range & $0.20 \sim 0.50$ \\
& Thickness    & $0.01 \sim 0.04$ \\

\midrule

\multirow{2}{*}{Triangle}
& Side range & $0.40 \sim 1.00$ \\
& Thickness & $0.01 \sim 0.04$ \\

\midrule

\multirow{3}{*}{Board\_X}
& Length range & $0.80 \sim 1.60$ \\
& Width range  & $0.04 \sim 0.20$ \\
& Thickness    & 0.30 \\

\midrule

\multirow{2}{*}{Tube\_X}
& Length range & $0.80 \sim 1.60$ \\
& Radius range & $0.01 \sim 0.08$ \\

\bottomrule

\end{tabular}

\end{table}
\subsection{Navigation Performance}
\label{sec:7.4}
\textbf{LOADNav Navigation Performance Across Payload Type.} Table~\ref{tab:7} reports the complete results summarized in Section~\ref{sec:4.2.1}. LOADNav and CoGE-LOADNav are evaluated on each of the six payload types and their mixture under obstacle densities $N=20$, $40$, $60$, and $80$, on a $5\times5$ terrain grid with 50 robots over 200 episodes per setting. The ACT of CoGE-LOADNav is higher because its estimate is refined through contact. Board\_Y and Tube\_Y, placed perpendicular to the heading, are the most challenging types; Board\_X and Tube\_X, aligned with the elongated body, are the easiest. 
\begin{table*}[h]
\centering
\scriptsize
\caption{Navigation Performance Across Payload Types and Obstacle Densities. ACT is averaged over all trials.}
\label{tab:7}

\setlength{\tabcolsep}{3.5pt}
\renewcommand{\arraystretch}{1.05}

\begin{adjustbox}{width=\textwidth}

\begin{tabular}{c | l | cc cc cc cc cc cc cc}

\toprule

\makecell{Numbers of\\Obstacles}
& \multicolumn{1}{c|}{Method}
& \multicolumn{2}{c}{Mix}
& \multicolumn{2}{c}{Board\_Y}
& \multicolumn{2}{c}{Tube\_Y}
& \multicolumn{2}{c}{Circle}
& \multicolumn{2}{c}{Triangle}
& \multicolumn{2}{c}{Board\_X}
& \multicolumn{2}{c}{Tube\_X} \\

\cmidrule(lr){3-4}
\cmidrule(lr){5-6}
\cmidrule(lr){7-8}
\cmidrule(lr){9-10}
\cmidrule(lr){11-12}
\cmidrule(lr){13-14}
\cmidrule(lr){15-16}

& \multicolumn{1}{c|}{}
& SR$\uparrow$ & ACT$\downarrow$
& SR$\uparrow$ & ACT$\downarrow$
& SR$\uparrow$ & ACT$\downarrow$
& SR$\uparrow$ & ACT$\downarrow$
& SR$\uparrow$ & ACT$\downarrow$
& SR$\uparrow$ & ACT$\downarrow$
& SR$\uparrow$ & ACT$\downarrow$ \\

\midrule

\multirow{2}{*}{N=20}
& LOADNav
& 98.50 & 0.015
& 97.50 & 0.035
& 97.01 & 0.035
& 99.50 & 0.005
& 97.00 & 0.015
& 98.00 & 0.030
& 99.01 & 0.015 \\

& CoGE-LOADNav
& 99.50 & 0.074
& 96.02 & 0.179
& 98.00 & 0.140
& 98.00 & 0.045
& 98.50 & 0.020
& 100.00 & 0.020
& 99.00 & 0.020 \\

\midrule

\multirow{2}{*}{N=40}
& LOADNav
& 93.60 & 0.059
& 92.54 & 0.090
& 93.00 & 0.070
& 95.50 & 0.035
& 95.50 & 0.045
& 95.00 & 0.060
& 94.50 & 0.035 \\

& CoGE-LOADNav
& 93.00 & 0.175
& 91.54 & 0.234
& 93.50 & 0.235
& 92.50 & 0.070
& 93.00 & 0.085
& 89.00 & 0.070
& 89.55 & 0.050 \\

\midrule

\multirow{2}{*}{N=60}
& LOADNav
& 87.00 & 0.155
& 88.62 & 0.158
& 88.50 & 0.150
& 88.50 & 0.060
& 90.50 & 0.080
& 91.63 & 0.118
& 90.10 & 0.069 \\

& CoGE-LOADNav
& 86.14 & 0.337
& 87.00 & 0.450
& 86.50 & 0.330
& 87.50 & 0.105
& 89.00 & 0.120
& 89.00 & 0.085
& 90.55 & 0.065 \\

\midrule

\multirow{2}{*}{N=80}
& LOADNav
& 80.50 & 0.160
& 81.09 & 0.234
& 84.50 & 0.230
& 81.50 & 0.170
& 85.57 & 0.114
& 85.57 & 0.139
& 85.50 & 0.130 \\

& CoGE-LOADNav
& 79.50 & 0.365
& 76.00 & 0.500
& 78.50 & 0.480
& 80.50 & 0.180
& 81.00 & 0.150
& 85.00 & 0.120
& 85.50 & 0.090 \\

\midrule
\multirow{2}{*}{\textbf{Average}}
& LOADNav
& \textbf{89.90} & \textbf{0.097}
& \textbf{89.94} & \textbf{0.129}
& \textbf{90.75} & \textbf{0.121}
& \textbf{91.25} & \textbf{0.080}
& \textbf{92.14} & \textbf{0.064}
& \textbf{92.55} & 0.089
& \textbf{92.28} & 0.062 \\
& CoGE-LOADNav
& 89.54 & 0.238
& 87.64 & 0.341
& 89.13 & 0.296
& 89.63 & 0.100
& 90.38 & 0.094
& 90.75 & \textbf{0.074}
& 91.15 & \textbf{0.056} \\

\bottomrule
\end{tabular}
\end{adjustbox}
\end{table*}

\begin{figure}[h]
    \centering
    \includegraphics[width=1.0\linewidth]{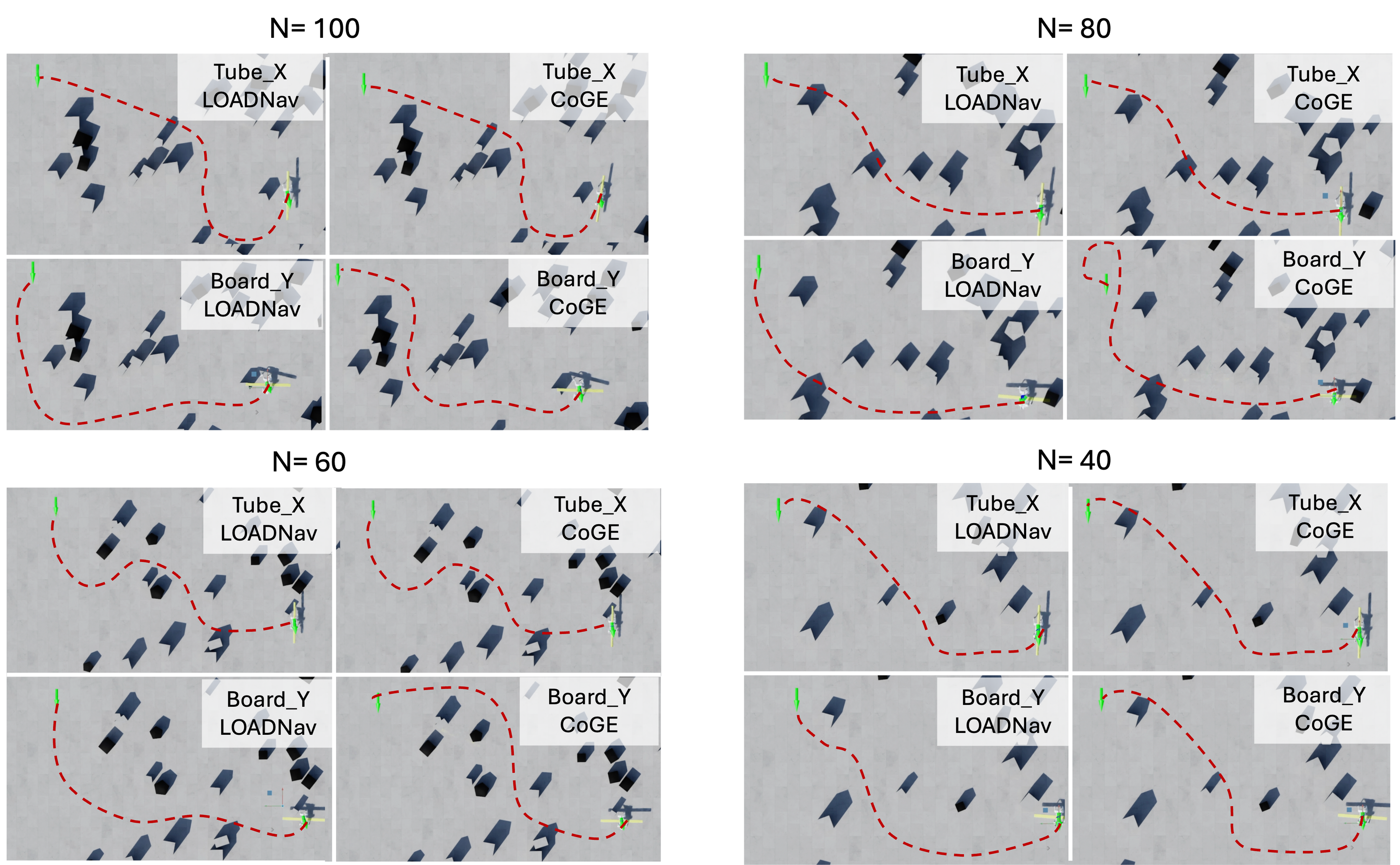}
    \caption{Simulation navigation trajectories of LOADNav and CoGE-LOADNav for Tube\_X and Board\_Y under varying obstacle densities (N = 40, 60, 80, 100).}
    \label{fig:9}
\end{figure}

\textbf{CoGE-LOADNav Navigation Performance.}
To further evaluate CoGE-LOADNav, we compare LOADNav and CoGE-LOADNav for Tube\_X and Board\_Y across varying obstacle densities in simulation, as shown in Fig.~\ref{fig:9}. For Tube\_X, CoGE-LOADNav produces trajectories nearly identical to those of LOADNav: since the payload is aligned with the robot's heading and adds little lateral extent, the robot navigates without collision and the predicted geometry remains stable throughout the episode. Comparing Board\_Y LOADNav with Tube\_X LOADNav, the load-aware navigation policy learned in Phase 1 consistently drives the robot to take wider detours when carrying an object with a larger Y dimension, demonstrating stable and geometry-aware path selection. Finally, for Board\_Y, CoGE-LOADNav starts without knowledge of the payload's lateral extent; once a collision occurs, the geometry estimate is corrected, which helps the robot escape and informs its subsequent navigation decisions.

\begin{figure}[h]
    \centering    \includegraphics[width=1.0\linewidth]{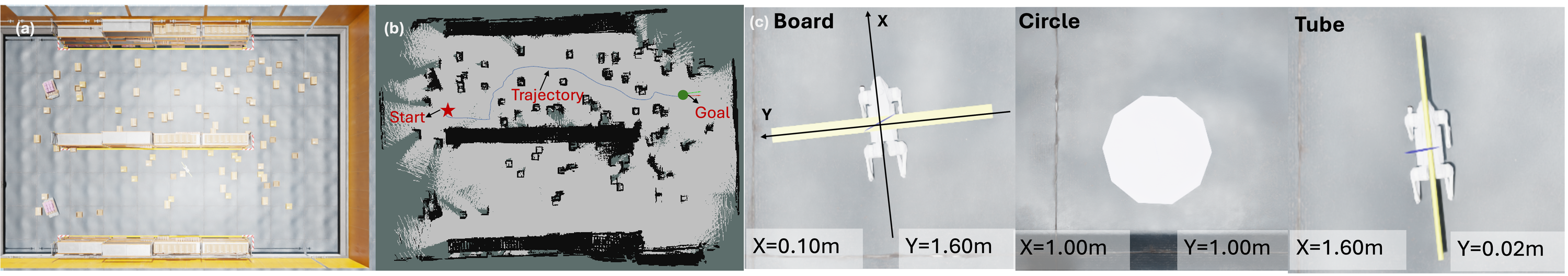}
    \caption{Warehouse benchmark environment. (a) Simulated warehouse with shelves and 100 randomly placed box obstacles, in which the Unitree Go2 is controlled through ROS2. (b) Occupancy map built online by RTAB-Map and a representative trajectory of LOADNav carrying a Circle payload. (c) The three payloads used in Table 1: Board\_Y (X = 0.10 m, Y = 1.60 m), Circle (X = 1.00 m, Y = 1.00 m), and Tube\_X (X = 1.60 m, Y = 0.02 m).}
    \label{fig:10}
\end{figure}

\textbf{Evaluation in the Simulation Benchmark.}
Fig.~\ref{fig:10} shows the benchmark environment used in Section~\ref{sec:4.4}. The warehouse contains shelves and 100 randomly placed box obstacles and communicates with the Unitree Go2 via ROS2. Occupancy maps are built online with RTAB-Map from the onboard RGB-D camera, and each method is evaluated over 20 trials per payload with the three payloads shown in Fig.~\ref{fig:10}(c).

\begin{figure}[h]
    \centering    \includegraphics[width=1.0\linewidth]{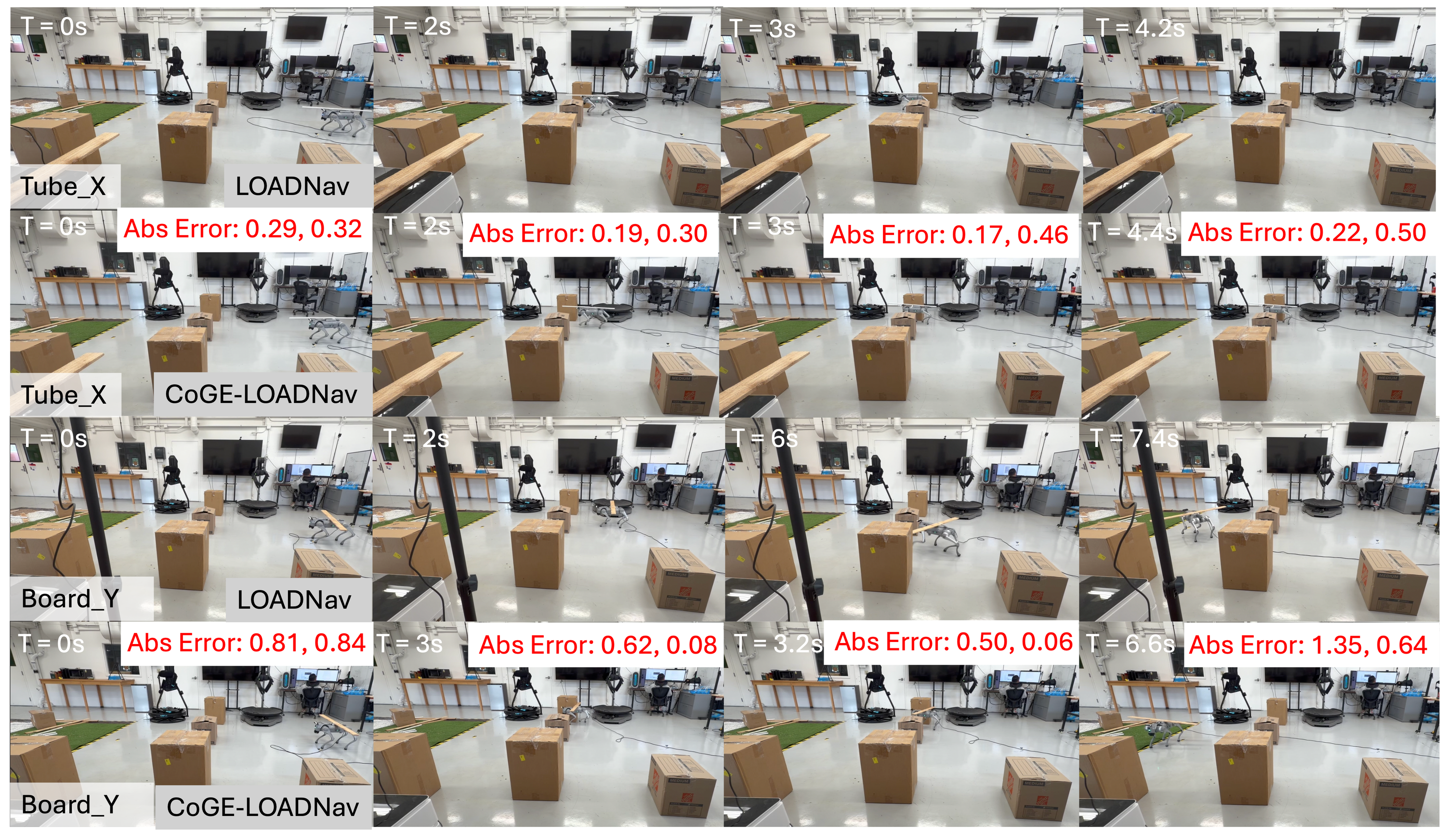}
    \caption{Real-world navigation examples of LOADNav and CoGE-LOADNav for Tube\_X and Board\_Y. }
    \label{fig:11}
\end{figure}

\textbf{Real-World Navigation Performance.}
Due to page limitations, we present more real-world examples here, as shown in Fig. \ref{fig:11}. For Tube\_X, both LOADNav and CoGE-LOADNav follow similar trajectories without any collision, and the absolute prediction error of CoGE-LOADNav remains stable throughout. For Board\_Y, LOADNav initially attempts to pass through a narrow space but pauses upon approach and reroutes to a wider passage. CoGE-LOADNav of Board\_Y encounters a collision in the narrow space, after which the robot actively rotates its body and successfully passes through to reach the goal. The absolute prediction error decreases significantly upon collision, reflecting improved geometry estimation through slight collision.
\subsection{Ablation}
\label{sec:7.5}
\begin{figure}[h]
    \centering    \includegraphics[width=1.0\linewidth]{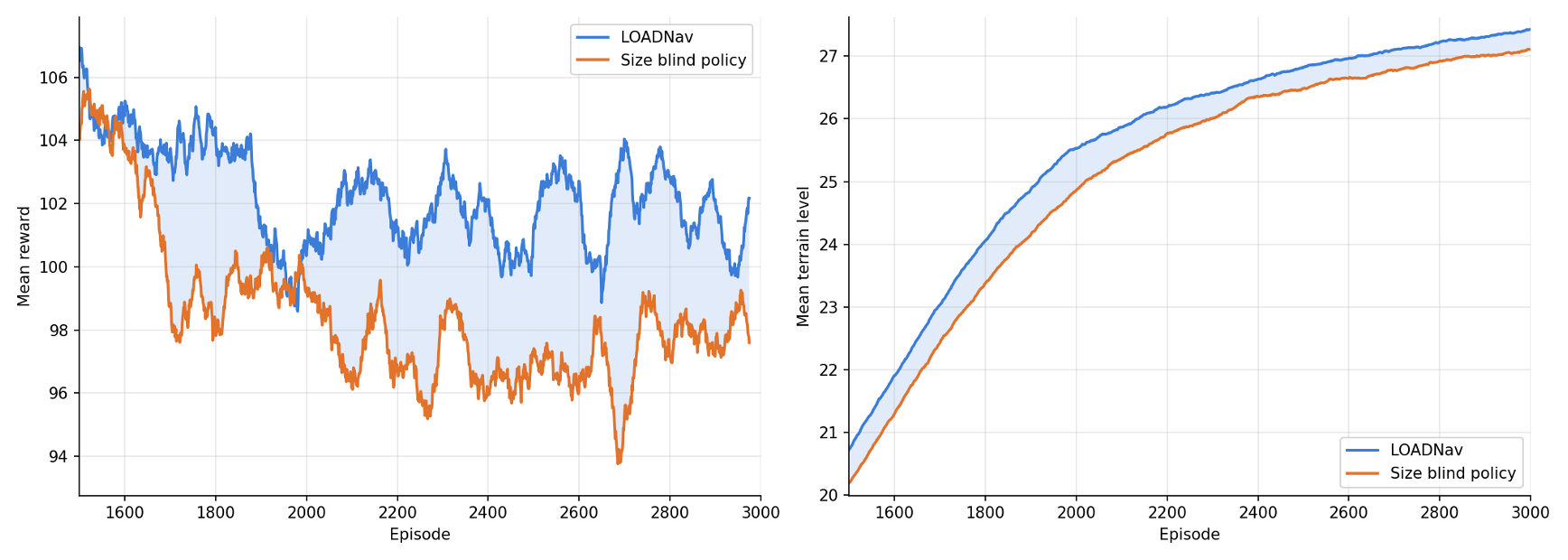}
    \caption{Phase 1 training curves of LOADNav and the LOADNav-B (size-blind policy) (episodes 1500–3000). Left: mean episode reward. Right: mean curriculum level, where a higher level corresponds to a denser obstacle field. Both policies share the curriculum, payload distribution, and PPO settings and differ only in the geometry input.}
    \label{fig:12}
\end{figure}

Fig.~\ref{fig:12} compares the Phase~1 training of LOADNav with that of LOADNav-B (the size-blind policy) used in Section~\ref{sec:4.4}. The mean reward decreases over training because the navigation curriculum raises the obstacle density as the policy improves. Throughout training, LOADNav attains both a higher reward and a higher curriculum level.
\end{document}